\pdfoutput=1

\documentclass[11pt]{article}

\usepackage[final]{acl}

\usepackage{times}
\usepackage{latexsym}

\usepackage[T1]{fontenc}

\usepackage[utf8]{inputenc}

\usepackage{microtype}

\usepackage{inconsolata}

\usepackage{graphicx}

\usepackage{color}
\usepackage{subcaption}
\usepackage{amsmath}
\usepackage{amssymb}
\usepackage{tabularx}
\usepackage{verbatim}
\usepackage{soul}
\usepackage{adjustbox}
\graphicspath{{imgs/}}
\usepackage[normalem]{ulem}

\title{Creative Problem Solving in Large Language and Vision Models -- What Would it Take?}

\author{Lakshmi Nair \\
  Georgia Institute of Technology \\
  Atlanta, GA, USA \\\And
  Evana Gizzi \\
  Tufts University \\
  Medford, MA, USA \\\And
  Jivko Sinapov \\
  Tufts University \\
  Medford, MA, USA
}

\begin{document}
\maketitle
\begin{abstract}
We advocate for a strong integration of Computational Creativity (CC) with research in large language and vision models (LLVMs) to address a key limitation of these models, i.e., creative problem solving. We present preliminary experiments showing how CC principles can be applied to address this limitation. Our goal is to foster discussions on creative problem solving in LLVMs and CC at prestigious ML venues.
\end{abstract}

\section{Introduction}
Creativity is ``\textit{...the ability to come up with an idea which, relative to the pre-existing domain-space in one's mind, one could not have had before. Whether any other person (or system) has already come up with it on an earlier occasion is irrelevant.}'' \cite{boden1998creativity}, p.216. For artificial agents, Computational Creativity (CC) is a multi-disciplinary field (spanning Philosophy, Psychology, Neuroscience, and Computer Science) that seeks to develop computational methods capable of generating creative outcomes reminiscent of creative processes in humans \cite{gizzi2022creative}. Within CC, \textit{creative problem solving} is a sub-area that requires an agent to discover -- from \textit{its} perspective -- novel and previously unseen ways to accomplish a task. For example, in the absence of a ladle to scoop ingredients, an agent might creatively choose to substitute a bowl in place of the ladle. In this sense, creative problem solving encompasses creativity that is \ul{specifically task-oriented}, as opposed to the generation of creative artifacts e.g., music or images.

\begin{figure}[t]
	\centering
\includegraphics[width=0.48\textwidth]{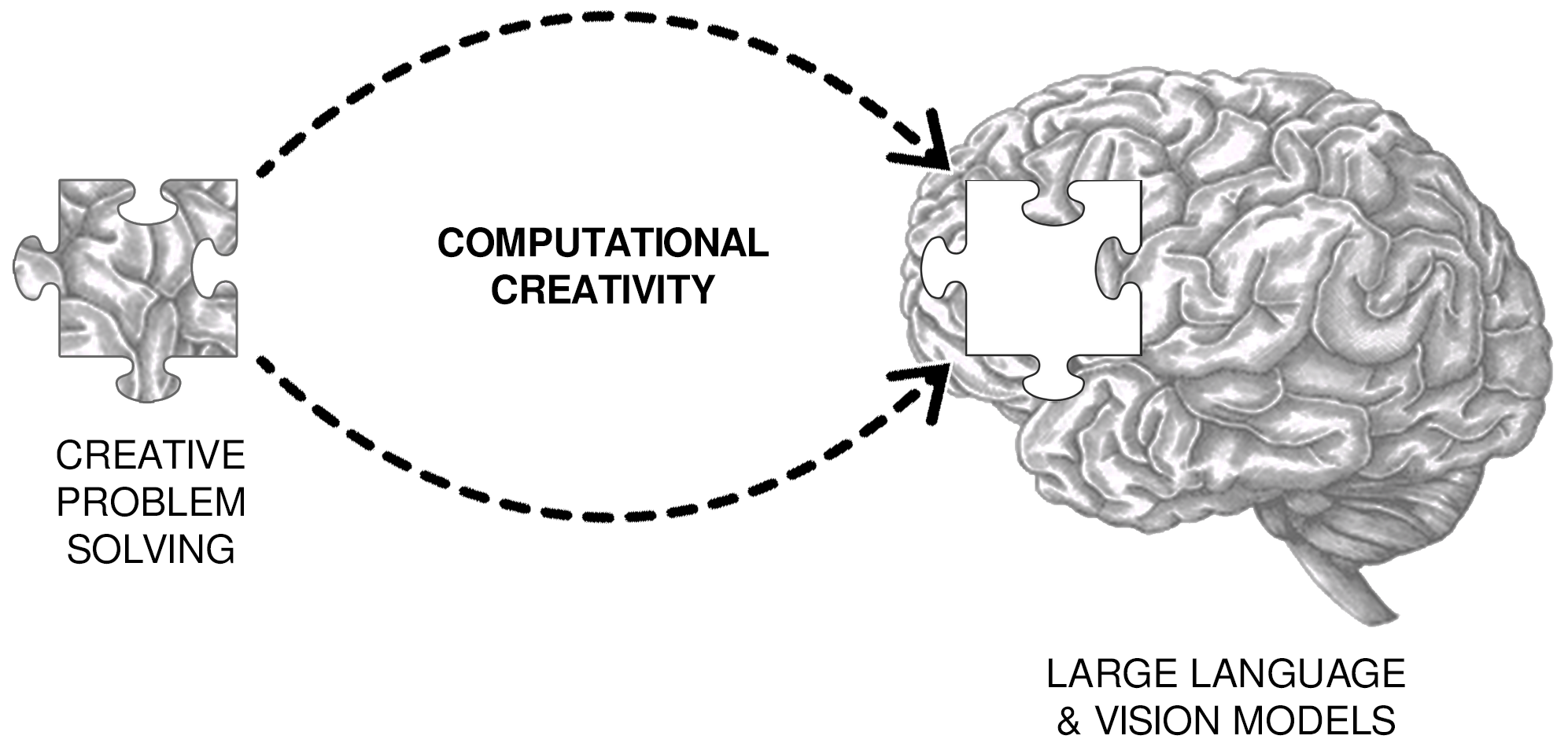}
	\captionsetup{width=\linewidth}
	\caption{Computational Creativity can help address a gap in the intelligence of present-day LLVMs, elevating their ingenuity through creative problem solving.}
	\label{fig:intro-img}
\end{figure}

While recent state-of-the-art large language models (LLMs) and vision-language models (VLMs) have demonstrated competency in artistic endeavours \cite{rombach2021high,copet2023simple}, creative problem solving continues to be a shortcoming of these models (we use LLVM to denote the umbrella of \textit{both} LLMs and VLMs). For instance, in \citet{bubeck2023sparks}, the authors point out that ``discontinuous tasks'' that require a certain ``Eureka'' idea, i.e., creative problem solving, is currently a limitation of models like GPT-4. Similar observations have been made in follow up work showing that state-of-the-art LLMs inherently possess poor creative problem solving capabilities compared to humans \cite{tian2023macgyver,naeini2023large}.

Given this obvious limitation, ongoing research in Machine Learning should seek to \ul{address the gap between LLVMs and creative problem solving, to further enhance the intelligent capabilities of these models}. We believe that a discussion of Computational Creativity is essential to addressing this limitation. It is our position that \textbf{Machine Learning and Computational Creativity should be strongly integrated in research to enable effective creative problem solving in LLVMs and push the frontiers of their ingenuity.} 

Creative problem solving can be a resourceful skill for artificial agents. As defined in prior work, ``\textit{Intelligence is the ability to work and adapt to the environment with insufficient knowledge and resources.}'' \cite{pennachin2007contemporary}, p.10. Demonstrated in hallmark examples of human ingenuity, like the makeshift $CO_2$ filter built onboard the Apollo-13 \cite{cass2005apollo}, or the makeshift medical devices used to offset equipment shortages during COVID-19 \cite{turner2020thinking}, creative problem solving is especially important when dealing with resource-critical scenarios. Since humans tend to ``choke'' under high pressure situations \cite{decaro2011choking} often limiting their problem solving skills, autonomous agents equipped with LLVMs that have similar capabilities would be highly assistive and transformative to humans in high-stake environments. These include situations like rescue missions \cite{bworld} or autonomous operation in human-inaccessible environments (e.g., space or underwater exploration) with limited resources \cite{atkeson2018happened}. However, the exceptional degree of creative problem solving necessary for such assistance remains beyond the scope of LLVMs today, limiting their intelligence (See Appx. \ref{subsec:agi_link}).




\section{Two Cultures Problem: Why does CC not receive a wider reception in ML?}
Even though creative problem solving (CPS) is a shortcoming of existing LLVMs, Computational Creativity seldom finds its way into mainstream ML research. We believe this discrepancy aligns with the ``two cultures'' problem \cite{hammond2013tale} (also corroborated in \citet{van2021ai,lahikainen2024creativity}), and is motivated by three aspects of CC literature as it relates to creative problem solving: a) the lack of a precise definition of CPS makes it challenging to identify how existing approaches in LLVMs are deficient in CPS skills; b) the somewhat ``abstract'' computational descriptions of CPS in Computational Creativity is challenging to connect to practical algorithms in LLVMs; and c) the lack of standardized benchmarks make it harder to evaluate LLVMs for CPS. In our discussions relating to a) in Section \ref{subsec:definition}, b) in Section \ref{sec:augmenting}, and c) Section \ref{section:evaluation}, \ul{we hope to address these gaps and encourage the ML community to think about how LLVMs can be augmented with creative problem solving skills \textit{through a deeper discussion of Computational Creativity.}}



To emphasize the applicability of principles from CC for creative problem solving in LLVMs, we discuss the seminal work of Margaret A. Boden from CC literature that introduces three forms of creativity, namely, ``\textit{exploratory}'', ``\textit{combinational}'', and ``\textit{transformational}'' \cite{boden1998creativity}. Prior work has discussed the extension of Boden's forms of creativity to creative problem solving in AI \cite{gizzi2022creative}, however, their work does not include recent advances in LLVMs nor how Boden's principles can be extended to specific approaches for LLVMs.





Ongoing discussions by leading ML experts like Dr. Shane Legg, co-founder of DeepMind, have suggested that ``search'' could help such models perform creative problem solving, quote, ``\textit{... these foundational models are world models of a kind, and to do really creative problem solving, you need to start searching}'' \cite{patel2023llmsneedsearch}. There has also been speculation that OpenAI's $Q^*$ search (described as a ``significant breakthrough'' in popular media) could be targeting a similar approach \cite{wang2023nextbig,tong2023reuters}. Interestingly, we note that ``search'' as described here, can be linked to Boden's proposed ``exploratory'' approach (Section \ref{subsubsec:exploratory}). However, in Section \ref{sec:augmenting}, we posit that ``combinational'' and ``transformational'' modes should be equally emphasized to achieve creative problem solving in LLVMs.

Although we choose to expand on Boden's work as the focal point to drive our arguments in the main paper, it is not the only theory in CC that is relevant to this discussion. For completeness, we elaborate on additional CC theories and their applicability to creative problem solving in LLVMs in Appx. \ref{sec:alt_theories}.


\section{From Task Planning to Creative Problem Solving}
Creative problem solving can be broadly described as the process through which agents \textit{discover} novel ways of accomplishing a task that was unsolvable prior to the discovery. Computationally, creative problem solving can be achieved through planning, learning, or hybrid approaches \cite{gizzi2022creative}. Following a review of the different definitions of creative problem solving that have been proposed (Appx. \ref{sec:alt_cps_def}), we believe the following most closely connects to existing formalisms in ML.

\subsection{Definition of Creative Problem Solving}
\label{subsec:definition}
\citet{gizzi2022creative} define the notion of a \textit{concept}, as a state (of the environment and/or agent) or action. More generally, the authors denote $C_X$ as the set of all concepts relating to $X$ ($X$ denotes environment states $S$ or actions $A$). Hence, $C_S$ denotes the set of all environmental states, and $C_A$ denotes the set of agent actions. Formally, the authors state their definition as (Page 7, \citep{gizzi2022creative}): 

\smallskip \noindent \textit{Given an un-achievable goal due to an insufficient conceptual space, CPS refers to the process by which the agent discovers a new conceptual space $C'_X \nsubseteq C_X$, such that $C'_X = f(C_X)$ is the result of applying some function $f$ on the current conceptual space, enabling the agent to solve the previously unsolvable task by using $C'_X$}.

As a simplified example, let us assume a robot that has a goal $G$ of transferring beans from a jar to a cooker: $G=$\{$in$(beans, cooker)\}. Here, the initial state is defined as $C_S=$\{$in$(beans, jar), $hasContainability$(spoon)\}. Let the actions be defined as $C_A=$\{$scoop$(beans, $X$, $loc_s$, $loc_d$)\}. Here, $X$ refers to an object that satisfies $hasContainability(\cdot)$, for example a spoon, to scoop beans from $loc_s$ to $loc_d$. If the robot has access to a spoon, the robot can use it to scoop the beans from the jar to the cooker to meet the goal. However, what if the robot did \textit{not} have a spoon, but had a \textit{glass} instead? By the definition of $C_S$, the agent is unaware that $hasContainability$(glass) is true, making the goal un-achievable. By our definition, creative problem solving is the process by which the agent uses some function $f(\cdot)$ to discover a new conceptual space: $f(C_S)=C'_S=C_S\mathop{\cup}$\{$hasContainability$\
(glass)\}. This would allow the agent to solve the previously unsolvable task by using the glass to scoop the beans instead.

In essence, CPS arises when the agent uses what it knows, to discover something new and the newly discovered knowledge is applied to solve a previously impossible task. Boden's three forms of creativity that we focus on in this paper, denote three plausible functions for $f(C_X)$. We revisit the notion of conceptual spaces in Section 3.


In the remainder of this section, we discuss how typical task planning is achieved with LLVMs. We divide the discussion into three subsections based on the level of task planning abstraction where LLVMs are applied: a) high-level task planning, b) low-level task planning, and c) hybrid task planning. While not exhaustive, our review is meant to offer a general insight into how LLVMs are used for task planning, to identify entry points for introducing creative problem solving capabilities.

\subsection{LLVMs for high-level task planning}
Approaches for high-level task planning often involve using LLVMs to identify high-level goals for accomplishing a task. Some approaches to task planning with LLMs often take a user input specifying the task, and generate high-level task plans for accomplishing it. These approaches often use LLMs as a form of ``knowledge base'', to extract actionable task plans from the models via appropriate prompting \cite{huang2022language}, further iterating over the generated task plan with repeated calls to the LLM as needed \cite{prasad2023adapt}.

In the context of Reinforcement Learning (RL), prior work has focused on using LLMs to suggest high-level goals for an RL agent \cite{du2023guiding}. Dubbed as ELLMs (Exploring with LLMs), an RL agent provides its current state to an LLM via a prompt, and receives a goal suggestion from the LLM that is then used to shape the reward and the agent exploration. Further work has extended this approach to incorporate the use of experience memory \cite{zhang2023large}. Existing approaches have also used LLMs to generate directed acyclic graphs composed of sub-goal states to aid the exploration of an RL agent \cite{shukla2023lgts}.

\subsection{LLVMs for low-level task planning}
Approaches for low-level task planning involve using LLMs to generate low-level code for performing a task. In contrast to high-level planning, where high-level goals and sub-goals are generated, these approaches use LLMs to directly generate low-level execution code via appropriate API calls \cite{liang2023code}. Other approaches have also investigated the capacity of LLMs to generate task plans via a low-level planning language such as PDDL \cite{silver2023generalized}, including iterating over the generated plan descriptions in case of errors \cite{guan2023leveraging}. In terms of low-level planning using VLMs, prior work has introduced an approach that uses a diffusion model to generate robot trajectories conditioned on language and the current visual state of the robot \cite{chen2023playfusion}.

\subsection{Hybrid high and low-level planning with LLVMs}
Hybrid approaches use LLVMs both for high-level goal generation as well as low-level planning. For instance, in \citet{li2023interactive}, user inputs are passed as LLM prompts to generate high-level plans. The high-level plans are then converted to low-level plans for robot execution via LLMs specialized for coding. Other approaches have used a high-level LLM planner, a VLM perceiver, and a low-level LLM planner for re-planning with both visual and language inputs \cite{skreta2024replan}.

\subsection{Summary} Given this overview, we see that LLVMs both at the high-level and low-level, can be modified to incorporate creative problem solving into task planning. For instance, the high-level task plans generated can encompass a novel substitution for a missing object, whereas the low-level task plan can generate an appropriate trajectory for creatively using the object. While the above approaches could, in principle, be studied within the framework of creative problem solving, that is not usually how the problem is formulated; there is a lack of paradigms for studying creative problem solving beyond just, \textit{``do you solve the problem or not?''}. Creative problem solving needs a fundamental rethinking of the typical problem formulations and approaches in ML. The next section is aimed at ways in which ML approaches in LLVMs can be reformulated from the perspective of CC.

\section{Augmenting LLVM embedding spaces for creative problem solving}
\label{sec:augmenting}
In this section, we discuss how principles from CC can be extended to LLVMs for creative problem solving. We begin with Boden's definition of ``conceptual spaces'' as ``\textit{[conceptual space] is the generative system that underlies the domain and defines a certain range of possibilities: chess moves, or molecular structures, or jazz melodies}'' \cite{boden2005whatiscreativity}, p.18 and ``\textit{... in short, any reasonably disciplined way of thinking}'' \cite{boden1998creativity}, p.214. By this definition, \ul{the embedding space of an LLVM describes its conceptual space or ``\textit{its way of thinking}''}. Some evidence for this also comes from existing work that introduces an approach for enabling LLMs to interpret continuous embedding spaces via natural language. Given an embedding vector representing an interpolation of different concepts, the model is able to interpret a text prompt in the context of the supplied embedding  \cite{tennenholtz2023demystifying}. The embedding thus determines the model's way of thinking. Hence, a discussion of enabling creative problem solving in LLVMs should target their embedding space. To this end, we explore two questions: a) \textit{how} can LLVM embedding spaces be augmented to achieve creative problem solving, and b) \textit{what} information should they be augmented with? Aligning with our original position, we show that CC literature can offer insights into these questions.

\subsection{\underline{How} can LLVM embedding spaces be augmented?}
In this section, we draw parallels between Boden's three forms of creativity and existing approaches in LLVMs. We further elaborate on how the three forms of creativity may enhance the potential of LLVMs to perform creative problem solving. We note that the ML approaches discussed in this section do not specifically perform creative problem solving. However, we discuss how they could potentially be extended to do so, by leveraging references from the CC literature.

\subsubsection{Exploratory Creativity}
\label{subsubsec:exploratory}
Exploratory approaches involve exploration within the conceptual or equivalently, the embedding space of the model, and most closely relates to ``search''. Note that the term ``exploration'' here differs from its usage in RL, instead referring to exploration through the model's \textit{embedding space}. Several existing approaches in the ML literature involve searching the \textit{output space} of LLMs with the goal of improving the performance of these models. The ``tree-of-thought'' model generates a ``tree'' of next possible LLM outputs, and searches through the states via Breadth-first or Depth-first search to reach the desired goal state, often guided by heuristics \cite{yao2023tree}. Numerous other approaches have built upon a similar strategy, such as using Monte-Carlo Tree Search (MCTS) \cite{zhou2023language,feng2023alphazero}, beam search \cite{zhang2023planning} or integrating pruning to remove sub-par candidates \cite{golovneva2023pathfinder}.

\textbf{Extension of exploratory creativity to LLVMs:} An important point to note here is that these approaches involve searching exclusively within the \textit{output} ``solution space'' of the LLMs rather than \textit{directly} operating in the \textit{embedding space} itself. In contrast to operating in the solution space of the LLM, exploratory approaches directly within the LLMs' embedding space would not be limited by what the LLM can generate as output -- ``\textit{Some exploration merely shows us the nature of the relevant conceptual space that we had not explicitly noticed before}'' \cite{boden2005whatiscreativity}, p.18. To effectively reveal the full extent of the conceptual space for creative problem solving, the approach should not be limited by the outputs the LLVM can generate. Rather, the generated (creative) outputs itself should be the result of heuristic or non-heuristic based search within the model's embedding space. However, to the best of our knowledge current approaches have not focused on LLVMs from this perspective, and have also not applied search to embedding spaces of Vision-LMs. Regardless, exploratory approaches are still limited by the dimensions of the model's embedding space. ``\textit{To overcome a limitation in the conceptual space, one must change it in some way}'' \cite{boden2005whatiscreativity}, p.18 - this leads us to combinational and transformational creativity.

\subsubsection{Combinational Creativity}
Combinational approaches involve combining two concepts to create something new - ``\textit{A novel combination of two familiar ideas is something
which did not happen before.}'' \cite{boden1998creativity}, p.213. We can broadly translate this to a function that takes in multiple concepts within an LLVM's embedding space to output a novel concept.

One way of extending this definition to LLVMs involves applying cross-attention layers. The attention operation is defined as \cite{vaswani2017attention}:
\[Attention(Q, K, V) = softmax(\frac{QK^T}{\sqrt{d_k}})V,\]
where, $Q$, $K$ and $V$ denote query, keys and values respectively, and $d_k$ denotes the dimensionality of the keys. Cross-attention involves passing $K$ and $V$ from a \textit{different} model, e.g., in Flamingo \cite{alayrac2022flamingo}, the keys and values represent visual input (from a separate vision encoder) and queries represent a language input. By applying cross attention in this manner, the embedding space of a model can be extended with capabilities of another model. In \citet{bansal2024llm} the authors show that using cross-attention layers can help augment an \textit{anchor} LLM with an \textit{augmenting} LLM's capabilities to perform a task that the anchor LLM was incapable of achieving before - hinting at some creative possibilities of this method. 

Other approaches in LLVMs, while using ``combinations'' in some way, do not conform to the notion of \textit{combinational creativity}. This includes, for instance, approaches that perform arithmetic combination of LLM weights to enhance the model performance \cite{matena2022merging, ilharco2022editing}. Or approaches that combine image and text embeddings via concatenation \cite{kim2021vilt} or a scaled dot product at the output \cite{radford2021learning}. While these approaches may be useful in imparting multi-modal capabilities, however, they do not lead to combinational creativity since the combination occurs \textit{external} to the models as opposed to within the model's embedding space.
 
\textbf{Extension of Combinational Creativity to LLVMs:} The ML approaches described here involve combining embedding spaces across models. Existing approaches have not looked at combining concepts \textit{within} the \textit{same} model's embedding space. The extension of combinational creativity to LLVMs is much more apparent in the sense of \textit{conceptual blending} \cite{fauconnier2003conceptual} for generation of creative artifacts, e.g., via blending of artistic styles. However, the extension of combinational creativity to creative problem solving is less obvious, and CC literature offers us further insights for making this connection. Typical conceptual blending corresponds to a form of ``aesthetic combination'', whereas creative problem solving would benefit from ``functional combinations'' \cite{chen2018computational}. Functional combination combines the functions (as opposed to aesthetic) of two components, e.g., a coin combined with pliers could function as a makeshift screwdriver. The authors extend this framework to a combination of two nouns with a ``base'' noun (e.g., ``pliers'') and ``additive'' noun (e.g., ``coin''). An interesting possibility stems from this notion: Can a combination of embeddings of the same LLVM, corresponding to ``base'' and ``additive'' nouns (perhaps with some prior denoting the task), enable the LLVM to generate creative combinations of objects for solving a task? This question remains unexplored, and points to a potential research direction for LLVMs inspired by CC.

\subsubsection{Transformational Creativity}
Transformational approaches involve transforming existing conceptual spaces to produce new ones. Transforming conceptual spaces can involve ``\textit{altering existing rules}'' \cite{boden1998creativity}, p.216. One way of transforming the embedding space of a model involves fine-tuning or training \cite{franceschelli2023creativity}. However, additional insight into transformational creative problem solving comes from prior work in CC, which describes creative problems as those with a poorly defined structure where a solution is not immediately apparent \cite{olteteanu2014two}. And in such cases, ``...\textit{ re-representation being the process which transforms an ill-structured problem into a well-structured one with direct inference to a problem solution}'' \cite{olteteanu2014two}, p.1. The notion of ``re-representing'' or ``redefining'' the problem can be best captured in the input prompts provided to an LLVM. This most closely connects to \textit{prompt engineering} and \textit{in-context learning} (ICL). 

Prompt engineering augments LLVMs with task-specific hints, called prompts, to adapt the LLVM to new tasks \cite{gu2023systematic}. Relatedly, in-context learning is a prompting method that provides the LLVM with instructions to solve a new task without requiring additional training. Previous work has shown that in-context learning and gradient-based optimization are equivalent \cite{von2023transformers}, thus connecting ICL to training or fine-tuning. 

\textbf{Extension of transformational creativity to LLVMs:} Task re-representations for creative problem solving, through prompting or ICL, have not been well explored within ML. Prompt engineering and ICL is a challenging task, since model performance is strongly dependent on the chosen prompts \cite{rubin2021learning}, further compounded by the fact that creative problems are inherently poorly defined \cite{olteteanu2014two}. However, useful insights can be derived from CC literature. For instance, regarding problems that require creatively re-purposing objects, the \textit{Object-replacement-object-composition} (OROC) framework \cite{oltecteanu2016object} illustrates re-representations of tasks, which can be translated into prompts. The paper defines three different types of creative tasks involving objects, and their task re-representations as (from \cite{oltecteanu2016object}, p.16):
\begin{enumerate}
    \item Replace an unfound object needed for a task with other objects present in the environment: \textit{``If I do not have an object X, which I would normally use because of its affordance\footnote{Affordance is defined as the relation between an agent, action and object, e.g., bowls have the ``contain'' affordance for humans.} $Af_X$ , what other object Y could I use, so that I can get a similar affordance, $Af_X \approx Af_Y$?}''
    \item Compose objects. ''\textit{If I do not have object X with affordance $Af_X$ , which objects $Y_1; Y_2; ... ; Y_n$, could I use to construct $X$ or an object $X'$ with an equivalent or similar affordance, $Af_X \approx Af_{X'}, Af_X \approx Af_{Y1} + Af_{Y2} + ... + Af_{Yn}$?}''
    \item Decompose objects. ``\textit{If I do not have an object $X$ with affordance $Af_X$ , which objects $Y_1; Y_2; ... ; Y_n$ which are components of object $Y$ could I use to obtain an object $Y'_i$ with an equivalent or similar affordance, $Af_X \approx Af_{Y'i}$?}''
\end{enumerate}
For task re-representation, affordances can refer to object properties that are relevant to the task, e.g., in some cases the shape may be relevant and in other cases, the material \cite{oltecteanu2016object}. Within LLVMs, the affordances $Af_X$ or $Af_Y$ can be defined via natural language or other modalities such as images. In the following section, we present preliminary experiments on using LLVMs for object replacement, with prompts that are inspired by the above task re-representations. However, an in-depth application of these re-representations as defined in CC to in-context learning in LLVMs remains unexplored.

\subsubsection{Summary}
In the previous sections, we drew parallels between Boden's three forms of creativity and approaches in LLVMs, further emphasizing how principles from CC can potentially help enable creative problem solving skills in these models.

\textbf{Integration with task planning:} Given the three methods, we see that transformational and combinational approaches may be especially aligned with LLVMs for high-level task planning. In contrast, exploratory methods may be suited to low-level planning, e.g., trajectory generation.

\textbf{Creative problem solving as a combination of the three methods:} An effective approach to creative problem solving may require all the three methods described in this section. While papers have explored chaining of LLMs within frameworks (often via prompts) \cite{ehud2022mrkl,zhan2023unleashing}, the individual LLMs themselves do not exhibit the characteristics described here. Existing frameworks in CC have shown that achieving creative problem solving would take a combination of all three methods, each of which is triggered in different contexts \cite{olteteanu2014two}. This presents potential opportunities for ML approaches that develop frameworks using multiple LLVMs, e.g., extending CC frameworks such as ``\textit{CreaCogs}'' \cite{oltecteanu2016object} can be highly beneficial for productive developments in ML.

\begin{table}[t]
\centering
\begin{tabular}{c|c}
\textbf{Model}  & \textbf{Acc. \% (no creativity)} \\ \hline
CLIP-B-32       & 100.0\%                                     \\
CLIP-B-16       & 92.0\%                                      \\
CLIP-L-14       & 98.0\%                                      \\
CLIP-H-14-laion & 98.0\%                                      \\
ViLT-B-32       & 68.0\% \\
LLaVA & 98.0\%
\end{tabular}
\caption{Accuracy of the models in predicting the nominal use of objects with no creativity involved.}
\label{table:nominal-perf}
\end{table}

\subsection{\underline{What} information should LLVM embeddings be augemented with?}
In the previous section, we discussed three methods for augmenting LLVM embedding spaces. In this section, we explore the question: ``What information should be targeted by the three methods when augmenting the embedding space for creative problem solving?''. In the previous section, we discussed this in the context of OROC. According to the OROC framework \cite{oltecteanu2016object}, information about object affordances could enable models to re-represent the task, such that the solution becomes evident. We propose a small experiment to validate whether the principles of transformational creativity of OROC are useful to LLVMs. We note that creativity can occur in various contexts, e.g., creatively solving a math problem or creatively playing a chess move, each of which would require different information. However, to facilitate the discussion in this paper, we focus our scope on tasks that require innovatively replacing missing objects (OROC Task \#1).

\textbf{Note on embeddings vs. concepts:} Our work connects ``conceptual spaces'' (or ``concepts'') as defined in Computational Creativity literature, to ``embedding spaces'' (or ``embeddings'') as defined in typical LM literature. We use ``concepts'' and ``embeddings'' interchangeably in this context. We make this connection to note that existing methods in Computational Creativity that operate on conceptual spaces translate to ML algorithms that operate on the LM’s embedding space. In this section, we connect the concept of ``affordances'' to the ``embeddings'' of the LLVMs in our experiments. Our goal is to show how the model can be prompted via an approach inspired by transformational creativity, to connect affordances of two seemingly distinct objects, e.g., a bowl and a spoon that appear distinct, but share the containability affordance.

\subsubsection{Experiment Setup}
\label{subsubsec:expt}
We create a simple experiment setup that tests the ``object replacement'' principle from OROC, where we create test sets composed of images of objects for replacing one of five core objects: ``Scoop'', ``Hammer'', ``Spatula'', ``Toothpick'', and ``Pliers''. We create two groups of tests: a) a nominal group where the actual object itself is available in each test set and requires no replacement (which serves as a form of baseline) and b) an object replacement group, where the nominal tool is missing, and a creative replacement object should be chosen. 

For each group, we create test sets with 4 objects each, chosen from a set of RGB images of 16 objects (Appendix Figure \ref{fig:test-objects}). We create 10 test sets per core object (total 50 samples per model). Each test set only includes one ground truth object, along with three other random objects that will not suit as an appropriate replacement. In the nominal group, the ground truth is the actual object itself. In the object replacement group, the replacements are chosen based on self-assessment of the authors as (core object $\xrightarrow{}$ replacement): ``Scoop'' $\xrightarrow{}$ ``Bowl''; ``Hammer'' $\xrightarrow{}$ ``Saucepan''; ``Spatula'' $\xrightarrow{}$ ``Knife''; ``Toothpick'' $\xrightarrow{}$ ``Safety pin''; ``Pliers'' $\xrightarrow{}$ ``Scissors''. For each test case, we pass the images in the test set along with a prompt. We record whether the ground-truth object image was chosen by the model for the prompt (i.e., assigned highest output probability)\footnote{CLIP generates probabilities that given images correspond to a text. ViLT and LLaVA respond with a text, and we assess if the model responded ``yes'' with a high probability for the ground truth.}.

The nominal group is subjected to one type of prompt: ``\textit{Can this object be used as a $\bigl \langle core \_ object \bigl \rangle ?$}''. In the object replacement group, each test case is subjected to four types of prompts:
\begin{enumerate}
    \item Baseline (regular) prompt: the same prompt as used in the nominal cases to obtain a baseline.
    \item Prompt prepended with affordance information: the prompt includes additional information about the desired object affordances specified as object features. 
    \item Prompt prepended with task information: the prompt includes additional information about the desired task.
    \item Prompt prepended with task and affordance information: the prompt includes additional information on the task and object affordance.
\end{enumerate}
Case \#2 aligns with task re-representations of OROC, and we explore cases \#3 and \#4 for comparison. We formulate our affordance prompts as brief versions of OROC's task re-representations. According to \citet{oltecteanu2016object} affordances can be defined using shape features, which we apply to the prompts here. The full set of prompts is shown in Appendix Table \ref{table:prompt-list}. The models that we explore include versions of CLIP \cite{radford2021learning}, LLaVA \cite{liu2024visual}, and ViLT \cite{kim2021vilt} obtained from HuggingFace. We use different model sizes (\ul{B}ase, \ul{L}arge, \ul{H}uge) and patch sizes (14, 16, 32). The open-source code for reproducing our experiment results (including our dataset and test cases) is available at: \url{https://github.com/lnairGT/creative-problem-solving-LLMs}. Appendix \ref{sec:expt_setting} includes more details on the experiments.

\subsubsection{Results}
\begin{figure}[t]
	\centering
	\includegraphics[width=0.49\textwidth]{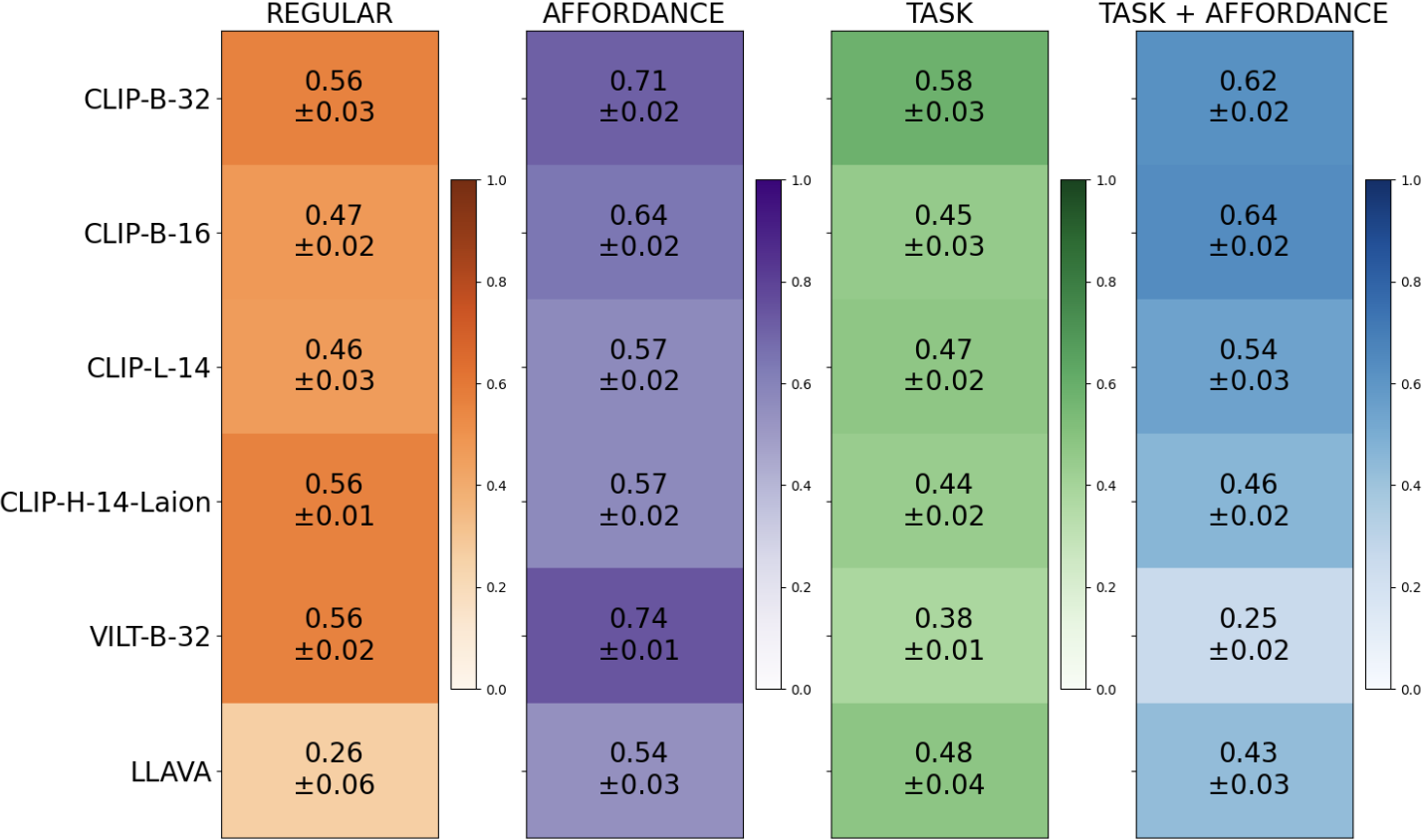}
	\captionsetup{width=\linewidth}
	\caption{Object replacement group: Average accuracies and standard deviations of the models across ten different sets of randomly chosen objects.}
	\label{fig:avg-perf}
\end{figure}

In Table \ref{table:nominal-perf}, we see the performances of the different models in the nominal test group, where the object requires no creative replacement. The models perform $>90\%$ in such cases (except for ViLT). In Figure \ref{fig:avg-perf}, we see the performances (accuracy shown on a $0.0-1.0$ scale) of the models in the object replacement test cases, where the object requires a creative replacement. For reference, a model that randomly picks an object achieves about 30\% overall accuracy. Figure \ref{fig:avg-perf} shows average accuracies for the different prompting strategies across random test sets. From Table \ref{table:nominal-perf} to Figure \ref{fig:avg-perf} (``regular''), the models perform poorly when they need to creatively reason about object replacements, highlighting their limitation. Comparing the ``Regular'' tab in Figure \ref{fig:avg-perf} to ``Affordance'', we see a \ul{general improvement in model performances, when object affordance information is provided}, consistent with description of the OROC framework \cite{oltecteanu2016object}. However, information about the task  (Figure \ref{fig:avg-perf}, ``Task'' ) leads to mostly detrimental results. Information about task \textit{and} affordances (Figure \ref{fig:avg-perf}, ``Task + Affordance'') does not lead to substantial improvements either, and is also detrimental in certain cases. We note that there is quite a variance in performances across the different models, which may be partially attributed to the original training datasets of the models. These observations warrant further exploration beyond the scope of this paper. Appendix \ref{sec:expt_results_more} includes a detailed, class-wise breakdown of the results.

\subsubsection{Summary}
While the experiments that we conducted are only preliminary, they offer some validity that the extension of principles in Computational Creativity can help overcome limitations of LLVMs in creative problem solving. The notion of task re-representation via improved prompting warrants further investigation in LLVMs, with regards to how the prompts can be generated automatically based on the creative task.

The models used in our experiments have all been trained jointly in visual and text domains. Multi-modal prompting capabilities may be useful for achieving creative problem solving. It can be quite challenging to describe affordances in words (example of ``hammers'' in our tests) and they may be better described through other means, e.g., images or depth maps or spectral data for material properties \cite{erickson2020multimodal}. This would require the application of multi-modal LLVMs that can process a variety of data types \cite{girdhar2023imagebind,han2023onellm}. Computational creativity can offer insight into meaningful representations of these different modalities that would help achieve creative problem solving, e.g., whether object material or shape matters more for one task vs. another \cite{oltecteanu2016object}. 

It is also worth noting that the creative problem solving examples in our experiments are human-centric. For instance, robots may not have similar capabilities as humans to manipulate bowls for scooping. In such cases, LLVMs need to account for the affordances as described \textit{with respect to the agent}, to derive creative solutions. However, that adds another level of complexity, yet to be explored, since these models are typically trained on human-centric data.

\section{Evaluation of Creativity}
\label{section:evaluation}
An important discussion in the context of creative problem solving is \textit{how can creative problem solving be evaluated?}. Prior work has proposed that creativity necessitates both \textit{novelty} and \textit{value} \cite{boden1998creativity,runco2012standard}, where the former guarantees that the generated outputs of a creative process are original, and the latter ensures that the generated outputs are useful. In the context of CPS, novelty refers to the discovery of new concepts (as defined in section \ref{subsec:definition}), while value insists that the newly discovered concepts successfully solve the task. Hence, CPS benchmarks should specifically evaluate \textit{how} the task was solved (novelty and value) rather than the typical ML evaluation of whether \textit{the task was successful or not} (value only). Some existing approaches that make this distinction describe problem settings that can be used to measure CPS skills of LLMs through the implicit integration of novelty and value measurements \cite{tian2023macgyver,naeini2023large,bisk2020piqa,talmor2022commonsenseqa}. In \citet{tian2023macgyver}, the authors create a dataset of 1600 real-world problems that necessarily involve creative reasoning abilities. Their proposed benchmark involves identifying novel approaches that can accomplish the given task (value). Similarly, in \citet{naeini2023large}, the authors introduce the Only-Connect-Wall (OCW) dataset to measure CPS capabilities of LLMs. The authors in \cite{bisk2020piqa} explore physical commonsense reasoning that is more generally applicable, beyond object-based creative problems. The authors introduce Physical Interaction: Question Answering, or PIQA consisting of 16,000 QA pairs where each question is paired with two possible common-sense solutions with a ground truth. In \cite{talmor2022commonsenseqa}, the authors introduce CommonSenseQA 2.0 (CSQA2) dataset consisting of both object-based and non-object based creative problems. The dataset consists of 14,343 questions distributed across 1,868 distinct topics. Currently, to the best of our knowledge, there are no standard benchmarks available to measure CPS skills of VLMs, although our preliminary experiments show one way to measure this using the task of object substitution.


\section{Conclusion and Future Work}
In this paper, we argued that an effective approach for enabling creative problem solving -- currently a key limitation of LLVMs -- should derive from Computational Creativity literature. To emphasize this at each juncture, we discussed the specific principles from CC that can be extended to achieve creative problem solving in LLVMs, describing the potential for further research with these insights.

It is rare to see special tracks or workshops targeted at Computational Creativity within more prestigious ML conferences. These programs typically focus on creative artifact generation and art (such as the NeurIPS \textit{Workshop on Machine Learning for Creativity and Design} \cite{NeurIPS2022creativity} or the recent tutorial at EMNLP on Creative Natural Language Generation \cite{chakrabarty2023creative}), but do not discuss CPS, thus failing to bridge the gap between CC and ML. We hope to see a deeper integration of the CC communities at such strong ML venues. We hope to encourage the reader to view creative problem solving and ML holistically, through the lens of Computational Creativity.


\section{Limitations}
\textit{Literature outside of Computational Creativity that enables CPS is unexplored:}
Our paper predominantly focuses on CC literature. This work does not cover literature beyond CC that can potentially inform creative problem solving in LLVMs. Although CC literature broadly encompasses psychology, neuroscience and philosophy, our future work seeks to explore specific literature within these sub-domains and discuss their applicability to creative problem solving and ML.

\noindent \textit{Lack of an explicit creative problem solving algorithm for LLVMs:}
Since the scope of our work aligns with a position paper, we have not focused on developing a concrete algorithm for creative problem solving in LLVMs. The prompting strategies explored in our preliminary experiments are manually specified, and our work does not elaborate on how these prompts may be automatically discovered. While our paper seeks to address some of the key gaps that prevent the application of CC literature to ML, there are still several unanswered questions when it comes to the practical implementation of an ML approach: e.g., what is a good representation for concepts that facilitate creative problem solving (symbolic, non-symbolic, or hybrid)? What is a good problem formulation for a given creative problem solving task (planning or learning)? etc. However, these questions are not directly answered within the scope of our work.

\section{Ethical Considerations}
The authors do not have specific ethical considerations to be highlighted with respect to this work.

\bibliography{latex/main}

\newpage
\appendix
\section{Alternate Definitions of Creative Problem Solving}
\label{sec:alt_cps_def}
Prior work by Olte{\c{t}}eanu \cite{olteteanu2014two} defines CPS from an object affordance perspective, where affordances broadly refer to action possibilities for objects, e.g., cups are pour-able and doors are open-able. The authors in \citet{olteteanu2014two} define creative problems as nominal problem solving tasks that have a poor representational structure, and as ``\textit{the ability of a cognitive, natural, or artificial system to use new objects to solve a problem, other than the ones that have been stored in
its memory as tools for that specific purpose (if any), or to create those objects by putting together objects or parts of objects the system has access to. Depending on the problem, objects can be either physical or abstract/informational (concepts, problem templates, heuristics or other forms of representations)}''. However, this definition is primarily object-creativity centered, and does not cover a wider range of creative problems.

Follow-up work by Sarathy and Scheutz \cite{sarathy2018macgyver}, define ``\textit{Macgyver-esque}'' creativity as a planning task that involves ``\textit{generating, executing, and learning strategies for identifying and solving seemingly unsolvable real-world problems}''. They introduce the ``\textit{MacGyver Problem}'' (MGP) as a planning problem with an unreachable goal state. Through the modification of the agent's domain knowledge (through \textit{domain expansion} and \textit{domain contraction}), the agent must discover new information and incorporate it into its existing domain knowledge, allowing the agent to accomplish the task. The domain expansion and contraction processes align with the divergent-convergent model of creative problem solving \cite{guilford1967creativity}. The definition of an MGP aligns well with the formulation of planning problems in ML, but less with learning or hybrid planning-learning approaches.

\section{Alternate theories on creative problem solving and their applications to ML}
\label{sec:alt_theories}
While there is exhaustive literature regarding theories on general creativity, we focus specifically on creative problem solving, with three well received works: \textbf{Divergent-Convergent Thinking} \cite{guilford1967creativity}, \textbf{Explicit-Implicit Interaction Theory} \cite{helie2010incubation}, and the \textbf{Creative Systems Framework} \cite{wiggins2006preliminary}. We discuss their applicability to ML in addition to the literature discussed in the main body of this paper to further emphasize the different pathways for connecting CC to LLVMs for creative problem solving. 


\subsubsection{Divergent-Convergent Thinking}
In \citet{guilford1967creativity}, the authors discuss the notion of ``divergent-convergent'' thinking. Divergent thinking or ``divergent-production'' (DP) abilities involve a more open-ended generation of a variety of ideas, whereas convergent thinking focuses on applying specific ideas to solve the problem.

\noindent \textbf{Applicability to CPS in LLVMs:} Prior work by \citet{tian2023macgyver} have demonstrated the applicability of ``divergent-convergent'' thinking towards solving \textit{Macgyver} problems. Similar in spirit to our experiments with VLMs in Section \ref{subsubsec:expt}, the authors prompt LLMs with descriptions of objects to enable the LLMs to reason about solving the task. Their work is, to the best of our knowledge, the only direct example demonstrating the value of CC literature in enabling CPS in LLMs.


\subsubsection{Explicit-Implicit Interaction Theory}
In \citet{helie2010incubation}, the authors introduce the \textit{Explicit-Implicit Interaction} (EII) theory, building upon the seminal work in \citet{wallas1926art}, that describes four stages of creativity: Preparation, incubation, illumination (i.e., insight), and verification. Preparation refers to the initial stage of searching in many different directions, which may fail to find a solution (i.e., impasse) in case of ill-defined problems (as is the case with CPS). Following an impasse, the incubation phase begins, where attention is not devoted to solving the problem. Over a period of time, illumination is the manifestation of the solution to the problem within the \textit{conscious} thought (i.e., \textit{``Aha''} moment). Finally, verification involves using deliberative thinking to assess if the solution indeed solves the problem.

\noindent \textbf{Applicability to CPS in LLVMs:} The authors in \cite{helie2010incubation} incorporate the four stages via a concrete computational method into the CLARION cognitive architecture. Prior work has also introduced a CPS framework for ML approaches inspired by the four stages \cite{gizzi2022creative}. In their work, ``preparation'' aligns with problem formulation, either task learning or planning. Incubation and illumination aligns with knowledge representation (symbolic, non-symbolic, or hybrid), and knowledge manipulation (functions that manipulate the conceptual space). Lastly, verification aligns with evaluation (via simulation, real-world platforms, or benchmarks). Although these works do not explicitly cover LLVMs and related algorithms, they demonstrate the value of integrating CC literature in ML, and can serve as useful starting points for ML approaches towards creative problem solving in LLVMs.

\subsubsection{Creative Systems Framework}
In \citet{wiggins2006preliminary}, the author expands on Boden's levels further in the context of a framework that formalizes creative systems. The paper defines: a) creative system, b) creative behavior, c) novelty, and d) value. The paper also discusses formalized notion of a \textit{universe of possibilities}, and \textit{conceptual spaces}. Crucially, the work describes the characteristics of a creative agent, that can help distinguish modes of \textit{failures} within a creative system, namely: a) \textit{hopeless uninspiration} -- where there are no valued concepts within the universe; b) \textit{conceptual uninspiration} -- where there are no valued concepts within the conceptual space of the agent; and c) \textit{generative uninspiration} -- where an agent is unable to find a valued concept owing to the specific method (e.g., search) employed.

\noindent \textbf{Applicability to CPS in LLVMs:} While the discussion of novelty, value and conceptual spaces in \citet{wiggins2006preliminary} aligns with our descriptions in Section \ref{sec:augmenting}, the different modes of uninspiration offers potential ways to assess failure modes in LLVMs. This allows agents to distinguish between systems where creative problem solving is not possible (hopeless uninspiration), as compared to systems where the conceptual space or the methodology for searching the conceptual space, may be at fault (conceptual or generative uninspiration). Although this approach has not been expanded in existing literature, it presents a promising direction for an evaluation framework that can distinguish CPS from non-CPS problems.

\subsection{A potential link between creative problem solving and general intelligence}
\label{subsec:agi_link}
Existing literature hints at a potential link between creative problem solving and Artificial General Intelligence (AGI) - systems that are broadly capable of solving almost all tasks that humans can \cite{shevlin2019limits}. For instance, in \citet{moruzzi2020artificial},  p.85., the author argues that there exists a strong correlation between creativity and AGI: ``\textit{... features that systems need to develop in order to achieve general intelligence are aspects that they need to possess also to earn the attribute creative}''. In \cite{goertzel2014artificial}, the author compiles a list of \textit{competencies} deemed essential for achieving AGI, including creative capacities like ``\textit{conceptual invention}'' and ``\textit{creative constructive play with objects}''. The processes of ``insight'' or ``incubation'' often associated with creative problem solving \cite{helie2010incubation,gilhooly2016incubation} is also considered important for AGI \cite{ventura2014can}. Taken together, it is likely that \ul{any promising vision of AGI would be incomplete without creative problem solving}.

Alongside the heavy ongoing discussion of AGI surrounding LLVMs \cite{bubeck2023sparks,fei2022towards,ma2023brain,xi2023rise,moor2023foundation,grudin2019chatbots}, there is often little to no discussion of creative problem solving or Computational Creativity within mainstream ML. As described in \citet{moruzzi2020artificial}, p.96, ``\textit{The investigation on the nature of creativity and on how it manifests itself not only in human but also in animal and artificial systems should, thus, not be intended as a niche discussion but, rather, as a fundamental research which can lay the foundations for further studies in artificial intelligence and its relation to humans}''. We hope that this work will encourage discussions of creative problem solving and Computational Creativity alongside discussions on AGI.

\section{Experiment Settings}
\label{sec:expt_setting}
\begin{table*}[]
\begin{tabular}{c|l}
\textbf{Prompt type}     & \multicolumn{1}{c}{\textbf{Prompt}}  \\ \hline
Regular       & \begin{tabular}[c]{@{}l@{}}``can this object be used as a scoop?''\\ ``can this object be used as a hammer?''\\ ``can this object be used as a spatula?''\\ ``can this object be used as a toothpick?''\\ ``can this object be used as pliers?''\end{tabular} \\ \hline
Affordance      & \begin{tabular}[c]{@{}l@{}}``scoops must be concave and hollow. can this object be used as a scoop?''\\ ``hammers must be heavy and have a handle attached to a cylinder at the end. \\can this object be used as a hammer?''\\ ``spatulas must have a handle attached to a flat surface at the end. \\can this object be used as a spatula?''\\ ``toothpicks must have a pointed tip. can this object be used as a toothpick?''\\ ``pliers must have two-prongs. can this object be used as pliers?''\end{tabular} \\ \hline
Task        & \begin{tabular}[c]{@{}l@{}}``scoops can transfer beans from one jar to another jar. can this object be \\used as a scoop?''\\ ``hammers can hit a nail into the wall. can this object be used as a hammer?''\\ ``spatulas can spread butter onto a pan. can this object be used as a spatula?''\\ ``toothpicks can pick food caught between the teeth. can this object be used\\as a toothpick?''\\ ``pliers can grab a coin. can this object be used as pliers?''\end{tabular} \\ \hline
Task and affordance & \begin{tabular}[c]{@{}l@{}}``scoops can transfer beans from one jar to another jar. scoops are concave\\and hollow. can this object be used as a scoop?''\\ ``hammers can hit a nail into the wall. hammers have a handle attached to a \\cylinder at the end. can this object be used as a hammer?''\\ ``spatulas can spread butter onto a pan. spatulas have a handle attached to a\\flat surface at the end. can this object be used as a spatula?''\\ ``toothpicks can pick food caught between the teeth. toothpicks have a\\pointed tip. can this object be used as a toothpick?''\\ ``pliers can grab a coin. pliers have two-prongs. can this object be used as\\pliers?''\end{tabular}
\end{tabular}
\caption{Prompts (across 4 groups) used in the experiment}
\label{table:prompt-list}
\end{table*}

\begin{figure*}[]
	\centering
	\includegraphics[width=0.95\textwidth]{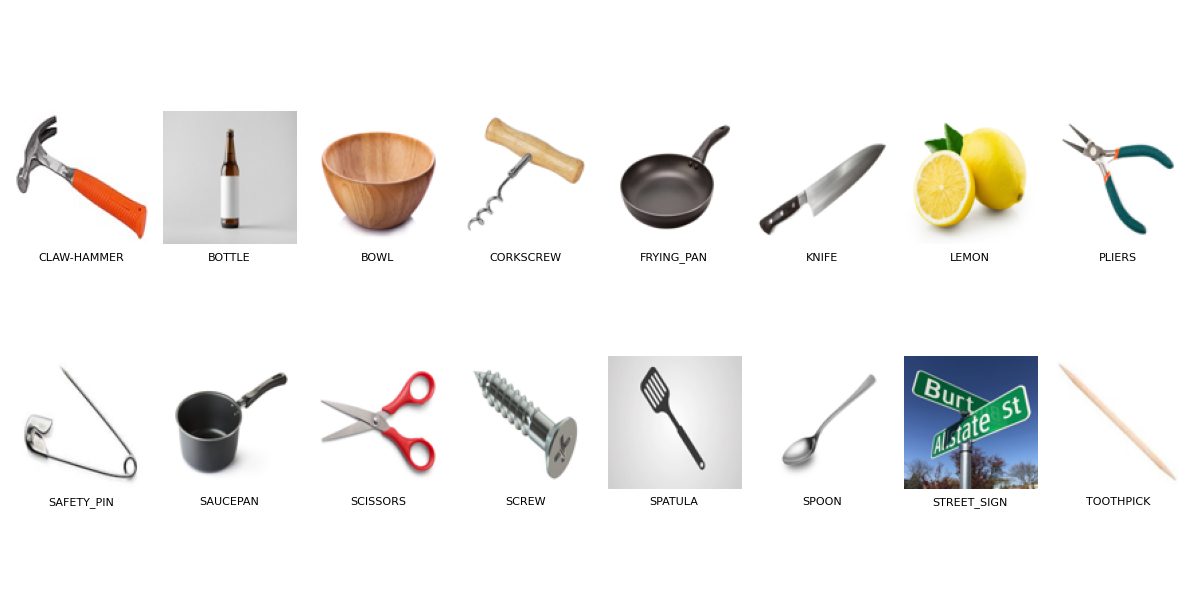}
	\captionsetup{width=\linewidth}
	\caption{Complete test set of objects used in the experiments.}
	\label{fig:test-objects}
\end{figure*}

\subsection{Data: Test images}
Figure \ref{fig:test-objects} shows the test set of 16 RGB images of objects used for the object substitution task. From the shown image dataset, we create test sets with 4 objects each, chosen from the set of 16 object images. We create 10 such test sets per core object (total 50 samples per model). Each test set only includes one ground truth object, along with three other random objects that will not suit as an appropriate replacement. In the nominal group, the ground truth is the actual object itself. In the object replacement group, the replacements are chosen based on self-assessment of the authors as (core object $\xrightarrow{}$ replacement): ``Scoop'' $\xrightarrow{}$ ``Bowl''; ``Hammer'' $\xrightarrow{}$ ``Saucepan''; ``Spatula'' $\xrightarrow{}$ ``Knife''; ``Toothpick'' $\xrightarrow{}$ ``Safety pin''; ``Pliers'' $\xrightarrow{}$ ``Scissors''.

\subsection{Model: Checkpoints}
For all the models, we use pre-trained HuggingFace checkpoints, with no additional training or fine-tuning. The models are of different architecture sizes and patch sizes: ``CLIP-B-32'' uses the \textit{``openai/clip-vit-base-patch32''} which is a base model with a patch size of 32. ``CLIP-B-16'' uses \textit{``openai/clip-vit-base-patch16''} -- a base model with patch size of 16. ``CLIP-L-14'' uses \textit{``openai/clip-vit-large-patch14''} -- a large model with patch size of 14. ``CLIP-H-14'' uses \textit{``laion/CLIP-ViT-H-14-laion2B-s32B-b79K''} which is a ``huge'' model, with a patch size of 14. This model is trained with the 2 billion sample English subset of LAION-5B. For LLaVA, we use the \textit{``llava-hf/llava-1.5-7b-hf''} with 7B parameters, version 1.5. Lastly, ``VILT-B-32'' uses \textit{``dandelin/vilt-b32-finetuned-vqa''} trained for visual question answering. However, there is limited data available on HuggingFace regarding the model.

\subsection{Prompts used in testing}
In this section, we discuss the prompts used in the different testing conditions (see Table \ref{table:prompt-list}). We explore four classes of prompts for the creative object substitution task: ``Regular'', ``Affordance'', ``Task'' and ``Task and affordance''. Regular prompts involve a direct prompt as to whether a given object will suffice as a substitute for the missing object. Affordance prompts, adds information about the desired affordances that are essential for replacing the missing object. Task prompts adds additional information on the task to be performed as context for whether a given object can be used as replacement for the missing object. Lastly, task and affordance prompts combine the task and object affordance information within the prompt.

\subsection{Testing Procedure}
For each test case, we pass the images in the test set along with a prompt belonging to one of the four classes described in Table \ref{table:prompt-list}. We record whether the ground truth object image was chosen by the model for the prompt (i.e., assigned highest output probability). CLIP generates probabilities that given images correspond to a text. ViLT responds with a text, and we evaluate if the model responded ``yes'' with a high probability for the ground truth.

\subsection{Testing Infrastructure}
We used NVIDIA-A100 GPUs to run the evaluation. However, the models are not too large and we have tested and confirmed that the code can be executed on CPU only as well.

\section{Continued Experiment Results}
\label{sec:expt_results_more}
\begin{figure}[h]
	\centering
	\includegraphics[width=0.48\textwidth]{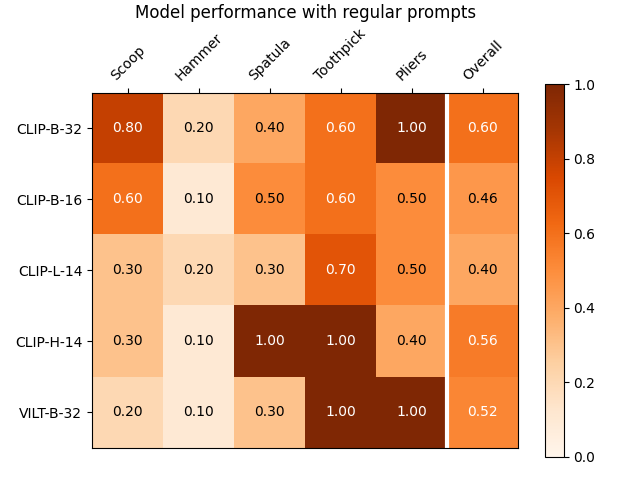}
	\captionsetup{width=\linewidth}
	\caption{Object replacement test: Using the same prompts  as for the nominal group. Random selection of a replacement object achieves $\approx$30\% overall accuracy.}
	\label{fig:viz-normal}
\end{figure}

\begin{figure}[h]
	\centering
	\includegraphics[width=0.48\textwidth]{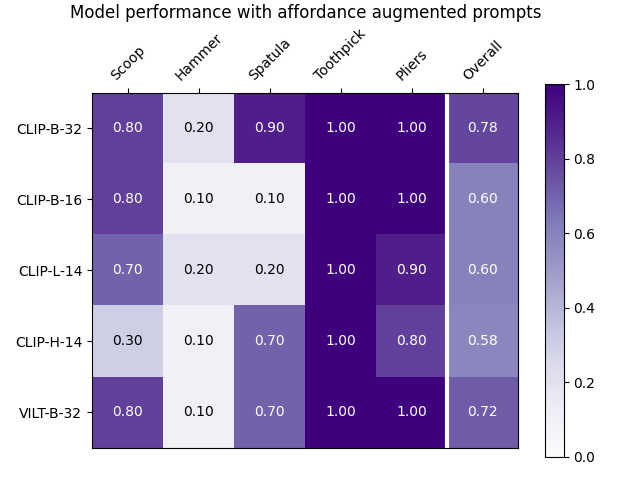}
	\captionsetup{width=\linewidth}
	\caption{Object replacement test: Accuracies when the prompts are augmented with object affordance information.}
	\label{fig:viz-obj}
\end{figure}

\begin{figure}[t]
	\centering
	\includegraphics[width=0.5\textwidth]{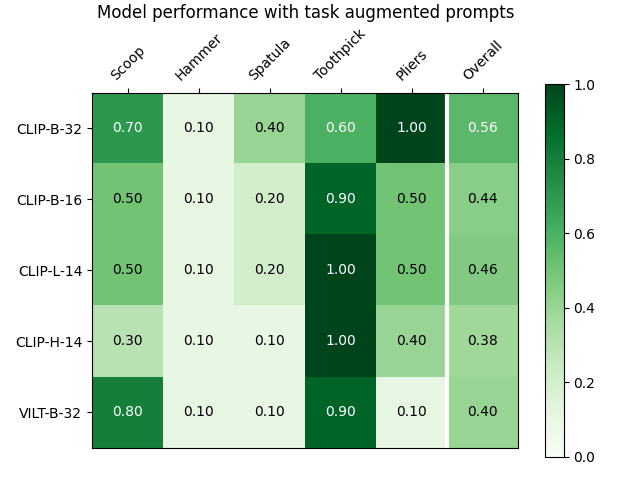}
	\captionsetup{width=\linewidth}
	\caption{Object replacement test: Accuracies when the prompts are augmented with task information.}
	\label{fig:viz-task}
\end{figure}

\begin{figure}[t]
	\centering
	\includegraphics[width=0.5\textwidth]{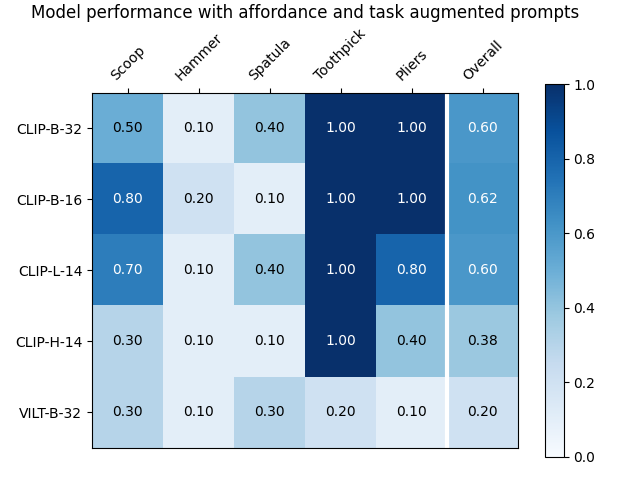}
	\captionsetup{width=\linewidth}
	\caption{Object replacement test: Accuracies when the prompts are augmented with task and object affordance.}
	\label{fig:viz-task-obj}
\end{figure}

In this section, we show the class-wise breakdown of the different models for the different prompting strategies (Figures \ref{fig:viz-normal} - \ref{fig:viz-task-obj}). We note that ``hammers'' present a particularly challenging case for all the models, perhaps due to the fact that correlating affordance of a hammer to a saucepan textually is difficult. In contrast, all models with the augmented prompts typically perform well in the case of creatively replacing ``toothpick'' with ``safety pin'' -- presumably indicating that specifying the relevant affordance textually in this case provides sufficient information. We repeated each experiment across multiple random seeds and found similar performances, showing that our general findings hold across different random cases. Generally, \ul{specifying object affordance information in the prompts leads to improved model performance.}

\end{document}